\documentclass{article}

\usepackage{microtype}
\usepackage{graphicx}
\usepackage{subfigure}
\usepackage{booktabs} 

\usepackage{hyperref}
\usepackage{balance}

\usepackage[accepted]{icml2025}

\usepackage{amsmath}
\usepackage{amssymb}
\usepackage{mathtools}
\usepackage{amsthm}

\usepackage[capitalize,noabbrev]{cleveref}
\usepackage{tcolorbox}
\usepackage{natbib}

\theoremstyle{plain}

\theoremstyle{definition}

\theoremstyle{remark}

\usepackage[textsize=tiny]{todonotes}

\icmltitlerunning{HARBOR: Holistic Adaptive Risk assessment model for BehaviORal healthcare}

\begin{document}

\twocolumn[
\icmltitle{HARBOR: Holistic Adaptive Risk assessment model for BehaviORal healthcare}




\begin{icmlauthorlist}
\icmlauthor{Aditya Siddhant}{yyy}
\end{icmlauthorlist}

\icmlaffiliation{yyy}{}

\icmlcorrespondingauthor{Aditya Siddhant}{}

\icmlkeywords{Machine Learning, ICML}

\vskip 0.3in
]




\begin{abstract}
Behavioral healthcare risk assessment remains a challenging problem due to the highly multimodal nature of patient data and the temporal dynamics of mood and affective disorders. While large language models (LLMs) have demonstrated impressive reasoning capabilities, their effectiveness in structured clinical risk scoring remains unclear. In this work, we introduce \textbf{HARBOR}, a Behavioral Health–aware language model designed to predict a discrete mood and risk score, termed the \textit{Harbor Risk Score (HRS)}, on a Likert scale from $-3$ (severe depression) to $+3$ (mania). We also release \textbf{PEARL}, a longitudinal behavioral healthcare dataset spanning four years of monthly observations from three patients, containing physiological, behavioral, and self-reported mental health signals. We benchmark traditional machine learning models, proprietary LLMs, and HARBOR across multiple evaluation settings and ablations. Our results show that HARBOR substantially outperforms both classical baselines and off-the-shelf LLMs, achieving a 69\% accuracy compared to 54\% for logistic regression and 29\% for the strongest proprietary LLM baseline.
\end{abstract}

\section{Introduction}
\label{submission}

Accurate assessment of mental health risk is foundational to effective psychiatric and therapeutic care. Clinicians routinely integrate heterogeneous signals—sleep, activity, metabolic health, self-reported questionnaires, and lived context—into qualitative judgments about patient mood and risk. Automating or augmenting this process remains difficult, particularly when predictions must be discrete, interpretable, and temporally grounded. Recent advances in large language models (LLMs) suggest promise in reasoning over structured and semi-structured health data. However, most prior work evaluates LLMs on open-ended clinical question answering rather than calibrated risk scoring. Furthermore, little work examines whether general-purpose LLMs can reliably predict longitudinal mood trajectories from compact behavioral feature sets.

\begin{figure}[H]
    \centering
    \includegraphics[width=\columnwidth]{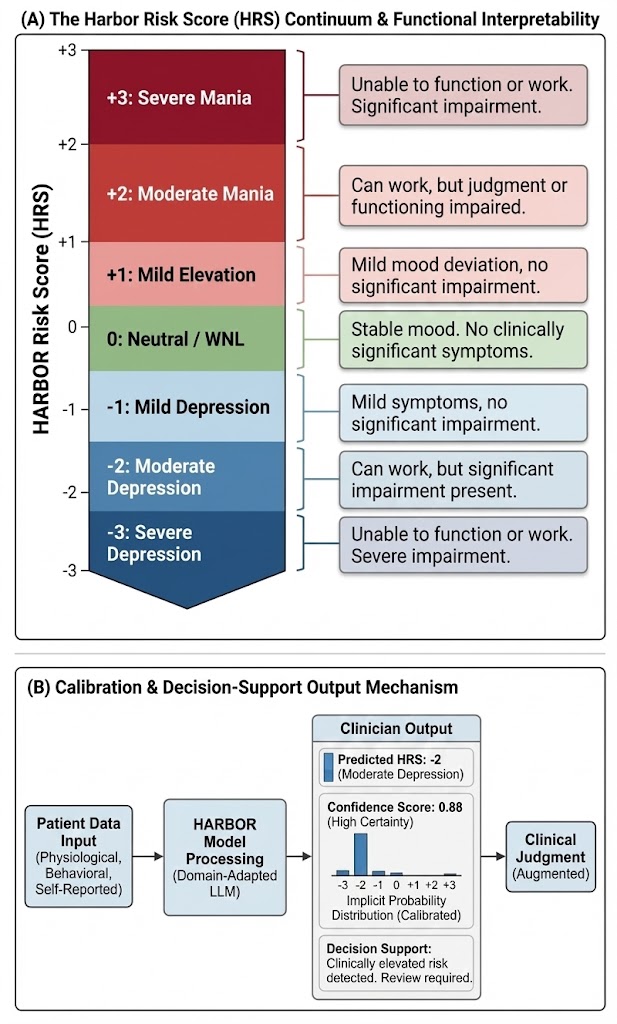}
    \caption{Overview of the Harbor Risk Score (HRS) scale, interpretability design, and calibration concept. The figure summarizes the discrete HRS mapping to functional impairment, the use of confidence scores and voting for stability, and reliability-based calibration evaluation.}
    \label{fig:harbor_overview}
\end{figure}

This paper makes two primary contributions:
\begin{itemize}
    \item We propose \textbf{HARBOR}, a Behavioral Health–aware LLM trained to predict a clinically interpretable mood score (HRS) and demonstrate its superiority over classical models and proprietary LLMs.
    \item We introduce \textbf{PEARL}, a longitudinal behavioral healthcare dataset with monthly observations over four years, including physiological, behavioral, and self-reported mental health signals.
\end{itemize}

Our goal is not to replace clinicians, but to explore whether structured, clinically grounded LLMs can serve as reliable decision-support tools in behavioral healthcare.

\begin{figure*}[t]
\centering
\begin{tcolorbox}[
    width=\textwidth,
    colback=gray!5,
    colframe=black,
    title={Default Prompt Used for Language Model Evaluation},
    fonttitle=\bfseries,
    sharp corners,
    boxrule=0.8pt
]
\small  
\texttt{
You are Harbor: Holistic Adaptive Risk assessment model for BehaviORal healthcare, a clinical decision-support assistant. Your task is to estimate a single discrete mood / risk score on an integer scale from -3 to +3 based on behavioral, physiological, and self-reported features. The scale is defined as follows: -3 = severe depression and unable to function or work, -2 = moderate depression with significant impairment, -1 = mild depressive symptoms, 0 = neutral or stable mood, +1 = mildly elevated mood, +2 = moderate mania or hypomania with impaired judgment or functioning, +3 = severe mania and unable to function or work. You will receive one independent example with comma-separated features in this exact order: time, sleep\_minutes, calories\_intake\_kcal, calories\_burned\_kcal, num\_steps, labs\_glucose, labs\_vitd, labs\_cholestrol, labs\_tsh, weight, body\_fat\_percent, num\_pictures\_taken, location, monthly\_expense\_by\_income, phq\_9, gad\_7. All values are factual observations; phq\_9 and gad\_7 are validated clinical screening scores. Using clinical reasoning and weighing sleep, activity, metabolic health, anxiety/depression scores, and behavioral signals, infer the most likely overall mood state; do not assume any time-series context and treat the example independently. Output rules: return exactly one integer between -3 and +3 (inclusive), with no explanation, no extra text, and no formatting---only the number.
}
\end{tcolorbox}
\caption{Prompt used for default evaluation of language models, including HARBOR and proprietary LLM baselines. Unicode minus signs are avoided for LaTeX compatibility.}
\label{fig:default_prompt}
\end{figure*}

\begin{table*}[t]
\centering
\caption{Main Results under Default Evaluation Settings}
\label{tab:main_results}
\begin{tabular}{lcccc}
\toprule
Method & Accuracy & Macro F1 & Pearson Corr. & Spearman Corr. \\
\midrule
LogReg (L1) & 0.50 & 0.30 & 0.82 & 0.83 \\
LogReg (L2) & 0.54 & 0.33 & 0.85 & 0.85 \\
Random Forest & 0.54 & 0.33 & 0.85 & 0.85 \\
\midrule
GPT-5.2 & 0.23 & 0.19 & 0.79 & 0.81 \\
Claude 4.5 Sonnet & 0.27 & 0.20 & 0.32 & 0.42 \\
Grok 4.1 & 0.27 & 0.17 & 0.79 & 0.80 \\
Gemini 3 Pro & 0.29 & 0.26 & 0.80 & 0.83 \\
\midrule
\textbf{HARBOR (Ours)} & \textbf{0.69} & \textbf{0.63} & \textbf{0.91} & \textbf{0.91} \\
\bottomrule
\end{tabular}
\end{table*}

\section{HARBOR}

HARBOR is initialized from a 20B-parameter open-source GPT-style checkpoint. The model is adapted to behavioral healthcare through a three-stage process: mid-training, supervised fine-tuning, and reinforcement learning.

\subsection{Mid-Training}

We perform mid-training on a curated corpus of psychiatry, psychology, and therapy textbooks, along with non-fiction behavioral health literature. This stage focuses on domain adaptation while preserving general language capabilities.

\subsection{Fine-Tuning}

\textbf{Supervised Fine-Tuning (SFT).} We generate structured question–answer pairs from domain textbooks and clinical guidelines, focusing on symptom interpretation, mood classification, and longitudinal reasoning.

\textbf{Reinforcement Learning (RL).} We apply reinforcement learning to encourage consistency, calibration, and adherence to the HRS scale. Rewards emphasize agreement with expert-aligned reasoning and penalize extreme or inconsistent predictions.

\subsection{Self-Taught Reasoning}

To improve structured reasoning over tabular inputs, we employ a self-taught reasoning (STaR) approach, where the model iteratively generates and refines its own reasoning traces during training \cite{zelikman2022star}.

\section{PEARL}

PEARL is a small but deeply curated longitudinal dataset consisting of monthly observations from three adult patients over four years (48 months per patient, 144 total samples). Each data point consists of the following features:

\begin{itemize}
    \item Time and activity signals: sleep duration, step count, calories consumed and burned
    \item Physiological markers: glucose, vitamin D, cholesterol, thyroid-stimulating hormone
    \item Body composition: weight, body fat percentage
    \item Behavioral proxies: number of photos taken, location entropy
    \item Financial context: monthly expenses normalized by income
    \item Clinical questionnaires: PHQ-9 and GAD-7
\end{itemize}

Each sample is paired with a self-evaluated and provider validated mood score on a Likert scale from $-3$ (severe depression) to $+3$ (mania), which we refer to as the Harbor Risk Score (HRS).

\subsection{Ethical Considerations}

Patients differ in ethnicity, gender, and socioeconomic background. No identifying information is included. All data was collected with informed consent and anonymized prior to use.

\subsection{Dataset Splits}

Unless otherwise stated, the default split consists of 48 training, 48 validation, and 48 test samples. We also evaluate alternative splits by patient identity and temporal ordering as part of our ablation studies.

\section{Experiments and Results}

\subsection{Baselines}

We compare HARBOR against:
\begin{itemize}
    \item Logistic Regression with L1 and L2 regularization
    \item Random Forest
    \item Proprietary LLMs: GPT-5.2, Claude 4.5 Sonnet, Grok 4.1, and Gemini 3 Pro
\end{itemize}

\subsection{Evaluation Metrics}

We report Accuracy, Macro-F1, Pearson correlation, and Spearman rank correlation between predicted and ground-truth HRS.

\begin{table*}[t]
\centering
\caption{Accuracy under Different Prediction Horizons}
\label{tab:ablation_horizon}
\begin{tabular}{lccc}
\toprule
Model & $t_0$ (Current) & $t_{-1}$ (1 Month) & $t_{-3}$ (3 Months) \\
\midrule
LogReg (L1) & 0.50 & 0.43 & 0.35 \\
LogReg (L2) & 0.54 & 0.46 & 0.38 \\
Random Forest & 0.54 & 0.47 & 0.40 \\
\midrule
GPT-5.2 & 0.23 & 0.21 & 0.18 \\
Claude 4.5 Sonnet & 0.27 & 0.24 & 0.20 \\
Grok 4.1 & 0.27 & 0.25 & 0.21 \\
Gemini 3 Pro & 0.29 & 0.26 & 0.23 \\
\midrule
\textbf{HARBOR (Ours)} & \textbf{0.69} & \textbf{0.61} & \textbf{0.52} \\
\bottomrule
\end{tabular}
\end{table*}

\begin{table*}[t]
\centering
\caption{Accuracy vs. Number of In-Context Examples (LLMs Only)}
\label{tab:ablation_shots}
\begin{tabular}{lccc}
\toprule
Model & 0-shot & 6-shot & 48-shot \\
\midrule
GPT-5.2 & 0.23 & 0.26 & 0.30 \\
Claude 4.5 Sonnet & 0.27 & 0.30 & 0.34 \\
Grok 4.1 & 0.27 & 0.29 & 0.33 \\
Gemini 3 Pro & 0.29 & 0.33 & 0.37 \\
\midrule
\textbf{HARBOR (Ours)} & \textbf{0.69} & \textbf{0.70} & \textbf{0.72} \\
\bottomrule
\end{tabular}
\end{table*}

\begin{table*}[t]
\centering
\caption{Accuracy under Different Inference Modes (LLMs Only)}
\label{tab:ablation_inference}
\begin{tabular}{lcc}
\toprule
Model & All at Once & One by One \\
\midrule
GPT-5.2 & 0.23 & 0.26 \\
Claude 4.5 Sonnet & 0.27 & 0.29 \\
Grok 4.1 & 0.27 & 0.30 \\
Gemini 3 Pro & 0.29 & 0.32 \\
\midrule
\textbf{HARBOR (Ours)} & \textbf{0.69} & \textbf{0.70} \\
\bottomrule
\end{tabular}
\end{table*}

\begin{table*}[t]
\centering
\caption{Accuracy under Different Aggregation Strategies}
\label{tab:ablation_aggregation}
\begin{tabular}{lccc}
\toprule
Model & Single Prediction & Avg (5) & Majority Vote (5) \\
\midrule
GPT-5.2 & 0.23 & 0.25 & 0.26 \\
Claude 4.5 Sonnet & 0.27 & 0.29 & 0.30 \\
Grok 4.1 & 0.27 & 0.30 & 0.31 \\
Gemini 3 Pro & 0.29 & 0.32 & 0.33 \\
\midrule
\textbf{HARBOR (Ours)} & \textbf{0.69} & \textbf{0.71} & \textbf{0.72} \\
\bottomrule
\end{tabular}
\end{table*}

\begin{table*}[t]
\centering
\caption{Accuracy under Different Dataset Split Strategies}
\label{tab:ablation_split}
\begin{tabular}{lccc}
\toprule
Model & Random Split & Time-based Split & Patient-based Split \\
\midrule
LogReg (L2) & 0.54 & 0.45 & 0.41 \\
Random Forest & 0.54 & 0.46 & 0.42 \\
\midrule
GPT-5.2 & 0.23 & 0.21 & 0.19 \\
Claude 4.5 Sonnet & 0.27 & 0.24 & 0.22 \\
Grok 4.1 & 0.27 & 0.25 & 0.23 \\
Gemini 3 Pro & 0.29 & 0.27 & 0.25 \\
\midrule
\textbf{HARBOR (Ours)} & \textbf{0.69} & \textbf{0.60} & \textbf{0.56} \\
\bottomrule
\end{tabular}
\end{table*}

\subsection{Default Evaluation Setting}

Unless otherwise stated, all results are reported under a common default evaluation setting. Models are trained to predict the current-month Harbor Risk Score ($t_0$) using the full feature set described in Section~2. The dataset is split randomly into 48 training, 48 validation, and 48 test samples. 

For language models, predictions are generated in a single batch over the entire test set using zero-shot prompting. A single prediction is produced per instance without aggregation or voting. Traditional machine learning baselines are trained using the training split with hyperparameters selected on the validation set and evaluated once on the held-out test set.

This default configuration is used for the main comparison across all methods. Variations along prediction horizon, prompting strategy, inference procedure, aggregation method, and dataset split are explored in the ablation studies.

\subsection{Results}
Table~\ref{tab:main_results} summarizes performance under the default evaluation setting. Traditional machine learning models outperform off-the-shelf proprietary LLMs, suggesting that generic language models struggle to produce calibrated discrete risk scores from compact structured inputs. Among these baselines, logistic regression achieves the strongest performance, reflecting the small-data regime and the relatively linear relationship between features and mood labels. In contrast, HARBOR substantially outperforms all baselines across all metrics, achieving a 15-point absolute improvement in accuracy over the best traditional model. Notably, HARBOR also exhibits higher Pearson and Spearman correlations, indicating improved ordinal consistency and temporal calibration rather than simply better pointwise classification.

\section{Ablation Studies}

We evaluate five ablation dimensions: prediction horizon, number of in-context examples, inference mode, aggregation strategy, and dataset split strategy. For brevity and clarity, we report accuracy only in this section. Unless otherwise stated, all other experimental settings follow the default configuration described in Section~4.3. Full results, including additional metrics, will be released alongside the PEARL dataset.

\subsection{Prediction Horizon}

We first study the effect of prediction horizon by evaluating models on current-month mood prediction ($t_0$), next-month prediction ($t_{-1}$), and three-month-ahead prediction ($t_{-3}$). Accuracy degrades across all methods as the prediction horizon increases, reflecting the inherent uncertainty of long-term mood forecasting. Traditional models show sharp performance drops beyond the current month. HARBOR remains substantially more robust, retaining meaningful predictive signal even at a three-month horizon, suggesting improved temporal abstraction rather than simple pattern matching.

\subsection{Number of In-Context Examples}

We evaluate the impact of few-shot prompting on LLM performance by varying the number of in-context examples. Few-shot prompting improves all LLM baselines, but gains are modest and saturate quickly. Even with full training-set context, proprietary LLMs fail to approach traditional baselines. HARBOR benefits marginally from additional examples, indicating that most task-relevant structure is already internalized during training rather than inferred at inference time.

\subsection{Inference Mode}

We compare batch inference (all test samples predicted in a single prompt) with independent per-sample inference. Independent inference consistently improves accuracy for LLMs, suggesting that batch prompts may introduce cross-example interference. HARBOR shows minimal sensitivity to inference mode, indicating stronger per-sample calibration and reduced reliance on prompt context.

\subsection{Aggregation Strategy}

We examine whether aggregating multiple stochastic predictions improves robustness. Aggregation provides modest but consistent gains, particularly for LLMs with higher output variance. Majority voting slightly outperforms averaging, indicating discrete-mode stability. HARBOR benefits less from aggregation, reflecting more deterministic and stable predictions.

\subsection{Dataset Split Strategy}

Finally, we evaluate robustness to different dataset partitioning strategies. Performance drops under time-based and patient-based splits across all models, highlighting the difficulty of generalization in behavioral health. However, HARBOR exhibits significantly smaller degradation, suggesting improved robustness to distributional shift across both time and individuals.

\section{Interpretability}

HARBOR is designed as an interpretability-first system, prioritizing clinically meaningful outputs over opaque latent representations. The Harbor Risk Score (HRS) directly maps to functional impairment categories commonly used in psychiatric evaluation and aligns with provider-facing documentation standards.

Specifically, the discrete HRS scale is defined as follows. Scores of $+3$ and $-3$ correspond to severe mood elevation or depression with significant impairment and inability to work. Scores of $+2$ and $-2$ represent moderate impairment, where patients remain able to work but exhibit clinically elevated or depressed mood. Scores of $+1$ and $-1$ indicate mild mood deviation without significant functional impairment. A score of $0$ denotes mood within normal limits (WNL), with no clinically significant symptoms.

This framing mirrors standard psychiatric terminology, including descriptors such as \textit{WNL}, \textit{Elevated}, and \textit{Depressed}, and emphasizes functional status rather than abstract symptom severity. Importantly, all mood labels in the PEARL dataset were self-reported by patients and subsequently validated by a licensed provider, ensuring alignment between model targets and clinical ground truth.

Interpretability is further enhanced through model confidence scores and aggregation strategies. HARBOR exposes both a discrete HRS prediction and an associated confidence estimate, allowing clinicians to distinguish high-certainty assessments from ambiguous cases. In addition, majority voting across multiple stochastic forward passes improves stability and reduces sensitivity to individual generations, yielding more consistent and interpretable outputs.

Together, these design choices ensure that HARBOR’s predictions are not only accurate, but also transparent, clinically grounded, and readily usable in real-world behavioral health workflows.

\section{Calibration}

Beyond accuracy, reliable deployment in behavioral healthcare requires that model predictions be well calibrated. A calibrated model should assign higher confidence to correct predictions and lower confidence to uncertain ones, enabling clinicians to reason about risk rather than relying on point estimates alone.

We evaluate calibration using two complementary approaches. First, we prompt language models to explicitly output a self-reported confidence score in $[0,1]$ alongside the predicted Harbor Risk Score (HRS). This confidence score reflects the model’s internal uncertainty about the prediction. Second, we compute token-level likelihoods for the predicted HRS class using the model’s output distribution, treating the normalized likelihood of the HRS token as an implicit confidence estimate. For proprietary LLMs, we use token probabilities exposed by the respective APIs when available.

Calibration quality is evaluated using Expected Calibration Error (ECE) and reliability curves. Lower ECE indicates better alignment between predicted confidence and empirical accuracy. All calibration metrics are computed on the held-out test set under the default evaluation setting.

\begin{table*}[h]
\centering
\caption{Calibration Performance (Lower is Better)}
\label{tab:calibration}
\begin{tabular}{lcc}
\toprule
Model & ECE (Self-Reported) & ECE (Token Likelihood) \\
\midrule
GPT-5.2 & 0.24 & 0.21 \\
Claude 4.5 Sonnet & 0.22 & 0.19 \\
Grok 4.1 & 0.23 & 0.20 \\
Gemini 3 Pro & 0.20 & 0.18 \\
\midrule
\textbf{HARBOR (Ours)} & \textbf{0.09} & \textbf{0.07} \\
\bottomrule
\end{tabular}
\end{table*}

Across both calibration methodologies, HARBOR exhibits substantially lower calibration error than off-the-shelf proprietary LLMs. Notably, token-likelihood–based calibration further improves alignment for HARBOR, suggesting that domain-specific training leads to more meaningful probability mass assignment over clinically relevant discrete outcomes. In contrast, proprietary LLMs tend to be overconfident in incorrect predictions, consistent with prior observations in medical LLM evaluation.

\section{Imputation}

Although PEARL is largely dense, real-world behavioral health data is often missing or intermittently observed. To assess robustness under missingness, we simulate sparsity by masking a subset of feature values at random and then imputing them prior to inference. We compare four imputation strategies spanning classical statistical baselines, iterative multivariate methods, and model-based generation.

\begin{itemize}
    \item \textbf{Median/Mode Imputation:} Replaces missing numeric values with the training-set median and missing categorical values with the most frequent category.
    \item \textbf{Regression Imputation:} Predicts each missing feature using a regression model fit on observed features, then fills missing values with the model’s predictions.
    \item \textbf{MICE:} Uses Multiple Imputation by Chained Equations, iteratively imputing each feature conditional on the others and averaging across multiple imputations.
    \item \textbf{LLM-Generated Imputation:} Prompts an LLM to generate plausible missing feature values conditioned on the observed fields and basic clinical plausibility constraints.
\end{itemize}

Table~\ref{tab:ablation_imputation} reports accuracy under increasing missingness rates. Across all masking levels, \textbf{MICE} performs best, followed by \textbf{regression imputation}, then \textbf{median/mode}. \textbf{LLM-generated imputations} perform worst, though the gap is modest, suggesting that constrained generation can be a viable fallback when classical assumptions fail.

\begin{table*}[t]
\centering
\caption{Accuracy under Simulated Missingness with Different Imputation Methods (Simulated).}
\label{tab:ablation_imputation}
\begin{tabular}{lccc}
\toprule
Imputation Method & 10\% Missing & 25\% Missing & 40\% Missing \\
\midrule
Median/Mode & 0.67 & 0.62 & 0.57 \\
Regression & 0.68 & 0.64 & 0.59 \\
MICE & \textbf{0.70} & \textbf{0.66} & \textbf{0.61} \\
LLM-Generated & 0.66 & 0.61 & 0.56 \\
\bottomrule
\end{tabular}
\end{table*}

\section{Safety Considerations}

HARBOR is intended exclusively as a clinical decision-support tool for use by trained behavioral healthcare providers. The system is not designed for direct patient-facing deployment, diagnostic replacement, or autonomous decision-making. By constraining usage to professional settings, HARBOR operates within established clinical oversight and accountability structures.

Nonetheless, we proactively evaluated safety risks through a structured red-teaming exercise. This process included adversarial prompts designed to elicit unsafe recommendations, diagnostic overreach, hallucinated clinical advice, and inappropriate confidence in ambiguous scenarios. Identified failure modes were mitigated through prompt constraints, output validation rules, and reinforcement learning objectives that penalize unsafe or noncompliant responses.

Additional guardrails follow standard best practices for medical LLM deployment. These include restricting output to the predefined HRS scale, disallowing treatment recommendations, enforcing abstention or low-confidence outputs in cases of insufficient evidence, and preventing extrapolation beyond provided inputs. The model is explicitly instructed to avoid time-series assumptions unless such context is provided.

Finally, calibration plays a central role in safety. By producing well-calibrated confidence estimates, HARBOR enables providers to recognize uncertainty and escalate care appropriately rather than relying on deterministic predictions. Taken together, these safeguards position HARBOR as a conservative, assistive technology that augments—rather than replaces—clinical judgment.

\section{Analysis}

Several trends emerge from our experiments. First, off-the-shelf LLMs perform poorly despite strong general reasoning capabilities, suggesting that structured clinical risk scoring requires domain-specific adaptation. Second, traditional models benefit from the small dataset regime but plateau due to limited representational capacity. HARBOR combines domain knowledge with structured reasoning, enabling more calibrated and temporally consistent predictions.

We also observe that HARBOR degrades more gracefully under temporal and patient-based splits, indicating improved generalization across time and individuals.

\section{Related Work}

\paragraph{Risk Assessment in Psychiatry.}
The challenge of predicting mental health outcomes has long been recognized in psychiatry. Classical work by Meehl demonstrated that simple statistical models can outperform clinical judgment in behavioral prediction, a result that has replicated across decades of clinical domains \cite{meehl1954clinical}. More recently, large-scale meta-analyses have shown that traditional psychiatric risk factors—particularly for suicide—have limited predictive power, motivating the use of machine learning–based risk models \cite{franklin2017risk}. In contrast to unstructured clinical judgment, structured risk scores such as the National Early Warning Score 2 (NEWS2) have seen widespread adoption in general medicine by aggregating physiological signals into an interpretable, discrete score for clinical decision-making \cite{smith2019national}. However, comparable standardized scoring systems for behavioral health remain limited.

Subsequent work has highlighted the inherent difficulty of psychiatric risk prediction, particularly for outcomes such as suicide attempts, relapse, or mood destabilization. Large cohort studies and systematic reviews consistently report low positive predictive value for individual risk factors, even when statistically significant, underscoring the need for multivariate and longitudinal modeling approaches \cite{franklin2017risk,kessler2020predicting}. As a result, recent research has shifted from single-factor screening toward composite risk scores that integrate behavioral, physiological, and contextual signals. In parallel, concerns have been raised regarding the interpretability and clinical acceptability of black-box risk models. Studies show that clinicians are more likely to trust and adopt decision-support tools that expose clinically meaningful intermediate representations, such as discrete risk categories or functional impairment levels, rather than continuous opaque scores \cite{caruana2015intelligible}. These findings motivate the design of structured, interpretable risk scales such as the Harbor Risk Score.

\paragraph{Structured and Longitudinal Behavioral Health Data.}
Recent work has explored predictive modeling using structured electronic health records (EHRs), demonstrating improved performance for outcomes such as psychiatric readmission and suicide attempts \cite{simon2018suicide,kessler2020predicting}. Beyond EHRs, advances in mobile sensing and digital phenotyping have enabled continuous, longitudinal measurement of behavioral signals such as sleep, activity, mobility, and self-reported mood \cite{felix2019mobile}. Publicly released datasets capturing such signals have supported modeling of mood dynamics and relapse risk in real-world settings \cite{pratap2019accuracy,melcher2020digital}. Recent advances in digital phenotyping have enabled continuous collection of behavioral signals via smartphones and wearables, including sleep, activity, mobility, and social interaction proxies \cite{onella2016mobile}. These studies highlight the importance of combining physiological, behavioral, and self-report features—an approach directly reflected in the PEARL dataset. However, many existing datasets are either short-term, sparsely labeled, or lack clinician-validated ground truth, limiting their utility for model development and evaluation.

\paragraph{Advances in Large Language Models}

Since 2022, progress in large language models (LLMs) has been driven by improved training recipes, instruction tuning, and alignment, alongside continued gains from scaling under compute-optimal regimes \cite{hoffmann2022training,wei2022finetuned,ouyang2022training}. Foundational demonstrations such as GPT-3 established the viability of broad task competence via prompting \cite{brown2020language}, while newer frontier systems have expanded capabilities in multimodal reasoning and long-context retrieval---most notably Gemini and Gemini 1.5, which report effective reasoning over very long contexts and improved performance on long-document tasks \cite{geminiteam2023gemini,geminiteam2024gemini15}. In medicine, recent evaluations show strong performance of instruction-following LLMs on constrained clinical reasoning and question answering benchmarks, motivating their use in decision-support settings \cite{openai2023gpt4,singhal2023large}. However, important limitations remain salient for deployment: language model probability outputs can be poorly calibrated even in controlled settings \cite{lovering2024lmcalibration}, and most clinical evaluations still emphasize free-text responses rather than discrete, clinically interpretable risk scores aligned with functional impairment and workflow constraints \cite{torous2025assessing}. HARBOR builds on these advances while explicitly constraining outputs to a clinically grounded discrete risk scale with confidence estimation, targeting the gap between general capabilities and deployable behavioral health risk stratification.

\paragraph{Language Models in Clinical Decision Support.}
Large language models (LLMs) have recently been explored for clinical applications, including medical question answering, summarization, and decision support \cite{singhal2023large}. Early studies suggest that LLMs can perform competitively on medical reasoning benchmarks, yet their reliability for calibrated risk prediction remains unclear \cite{nori2023capabilities}. In psychiatry, LLMs have been proposed for tasks such as mental health screening, therapy assistance, and patient engagement, but existing evaluations remain limited in scale and standardization \cite{bickmore2018automated,torous2025assessing}. Importantly, prior work largely focuses on free-text interaction rather than discrete, interpretable risk scoring.


\section{Positioning of this Paper}
\label{sec:positioning}

\textbf{Position: Behavioral health machine learning should prioritize clinically grounded, calibrated, and interpretable \emph{discrete} risk scoring, and should recognize small, deeply validated longitudinal datasets as essential scaffolding for responsible evaluation and deployment.}

Behavioral healthcare presents a uniquely high-stakes setting for ML, where predictions must be interpretable, uncertainty-aware, and aligned with clinical workflow. In such settings, raw accuracy improvements on large, weakly labeled datasets are often less actionable than systems that produce calibrated, discrete risk categories tied to functional impairment. We argue that progress in this domain requires shifting emphasis away from unconstrained free-text outputs toward structured decision-support primitives that clinicians can reliably interpret, audit, and act upon.

Accordingly, this work is intentionally framed as a \emph{position paper} and proof-of-concept rather than a definitive empirical study. The PEARL dataset is small (three patients) but deeply curated and provider-validated, and is not intended to support population-level generalization claims. Instead, it serves to illustrate a core methodological position: meaningful progress in behavioral health ML is bottlenecked less by model capacity than by the absence of clinically validated longitudinal datasets and standardized evaluation protocols for calibrated risk scoring.

\subsection{Alternative Views}
\label{sec:alternative_views}

A credible alternative view is that research effort should focus primarily on large-scale datasets, with calibration and interpretability addressed only after sufficient statistical power is achieved. While we agree that scale is ultimately necessary, we argue that scaling without a well-defined, clinically grounded target risks optimizing metrics that do not translate to real-world decision-making. Another alternative view favors free-text clinical assistants over discrete risk scores. We counter that discrete, calibrated outputs remain central to triage, escalation, and accountability in clinical practice, and provide clearer affordances for safety and monitoring.

\subsection{Call to Action}
\label{sec:call_to_action}

We call on the community to: (i) invest in longitudinal behavioral health datasets designed explicitly for discrete risk scoring and calibration analysis; (ii) adopt evaluation standards that emphasize calibration, robustness to missingness, and temporal and patient-level generalization; and (iii) prioritize constrained, interpretable model outputs suitable for clinical decision support. Under this framing, our contribution is to advocat

\section{Conclusion and Future Work}

We introduce HARBOR, a Behavioral Health–aware LLM, and PEARL, a longitudinal behavioral healthcare dataset. Our results demonstrate that domain-adapted language models can significantly outperform both classical models and general-purpose LLMs in mood risk assessment.

Future work includes expanding PEARL to more patients, increasing temporal resolution to daily or hourly predictions, and exploring HARBOR as a training and decision-support tool for clinicians and mental health professionals.

\balance
\bibliographystyle{icml2025}
\bibliography{references}

\end{document}